\documentclass[letterpaper]{article}
\usepackage{spconf,mathtools,graphicx,subfigure,tabu,booktabs,IEEEtrantools}
\graphicspath{{./figures/}}
\usepackage{flushend}
\usepackage[bookmarks=false,draft]{hyperref}

\title{Accelerating Discrete Wavelet Transforms on GPUs}

\name{
	David Barina and Michal Kula and Michal Matysek and Pavel Zemcik%
	\thanks{
		This work has been supported by
		the Ministry of Education, Youth and Sports of the Czech Republic from the National Programme of Sustainability (NPU II) project IT4Innovations excellence in science (LQ1602), and
		the Technology Agency of the Czech Republic (TA CR) Competence Centres project V3C -- Visual Computing Competence Center (no. TE01020415).
	}
}%
\address{Faculty of Information Technology\\%
	Brno University of Technology\\%
	Czech Republic%
}

\newcommand*\Figure[1]{Fig.~\ref{#1}}
\newcommand*\Table[1]{Table~\ref{#1}}

\let\P\undefined
\newcommand{\U}[1][]{{\mathrm{U}_{#1}}}
\newcommand{\P}[1][]{{\mathrm{P}_{#1}}}

\newcommand\even[1]{{#1}^{\!(e)}}
\newcommand\odd[1]{{#1}^{\!(o)}}

\def\ww{2.8em}
\newcommand\w[1]{\makebox[\ww]{$#1$}}

\newcommand\T[1][]{\mathrm{T}_{#1}}
\let\S\undefined
\newcommand\S[1][]{\mathrm{S}_{#1}}

\newcommand\NN[1][]{\mathrm{\mathbf{N}}_{#1}}

\begin{document}
\bstctlcite{IEEEexample:BSTcontrol}%
\ninept

\maketitle

\begin{abstract}
The two-dimensional discrete wavelet transform has a huge number of applications in image-processing techniques.
Until now, several papers compared the performance of such transform on graphics processing units (GPUs).
However, all of them only dealt with lifting and convolution computation schemes.
In this paper, we show that corresponding horizontal and vertical lifting parts of the lifting scheme can be merged into non-separable lifting units, which halves the number of steps.
We also discuss an optimization strategy leading to a reduction in the number of arithmetic operations.
The schemes were assessed using the OpenCL and pixel shaders.
The proposed non-separable lifting scheme outperforms the existing schemes in many cases, irrespective of its higher complexity.
\end{abstract}


\section{Introduction}
\label{sec:intro}

The discrete wavelet transform (DWT) has become a very popular image-processing tool in recent decades.
This popularity has resulted in a development of fast algorithms on all sorts of computer systems, including graphics processing units (GPUs).
Although the GPUs were originally optimized for the graphics rendering, general-purpose computing became practical with advances of programmable shaders.
Later, CUDA and related general-purpose computing APIs were introduced, which brought wider possibilities and allowed to ignore the underlying graphical concepts.

So far, several studies have compared the performance of various \mbox{2-D} DWT computational approaches on GPUs.
All of these works are based on the most popular separable schemes (their operations are oriented horizontally or vertically) -- the convolution and lifting schemes.
The lifting requires fewer arithmetic operations as compared with the convolution, at the cost of introducing some data dependencies.
The number of operations should be proportional to a transform performance.
Interestingly, also the data dependencies may form a bottleneck, especially on parallel architectures.

We show that the optimal scheme for a given architecture can be obtained by fusing the corresponding lifting parts into a new non-separable structure.
More precisely, underlying operations cannot be associated with horizontal nor vertical axes.
In addition, we discuss an approach where this non-separable scheme can be adapted to a particular platform in order to reduce the number of operations.
Our reasoning is supported by experiments on GPUs using OpenCL and pixel shaders.
The presented scheme is general, and it is not limited to any specific type of DWT.

The rest of this paper is organized as follows.
Section \ref{sec:background-and-related-work} formally introduces the problem definition and presents the existing separable approaches.
Section \ref{sec:proposed-schemes-and-improvements} presents the proposed non-separable scheme and discusses the optimization approach that reduces the number of operations.
Section \ref{sec:performance} evaluates the performance in the pixel shaders and OpenCL framework.
Finally, Section \ref{sec:conclusion} closes the paper.

\section{Background and Related Work}
\label{sec:background-and-related-work}

This section introduces some notations and definitions and also briefly reviews papers that motivated our research.
The $z$-transform notation is used for the description of underlying one-dimensional wavelet FIR filters.
The transfer function of the filter $\left( g_k \right)$ is the polynomial
\begin{align*}
	G(z) = \sum_{k} \, g_k \, z^{-k} \text{,}
\end{align*}
where the $k$ refers to the time axis.
Further, the one-dimensional transforms are used in conjunction with two-dimensional signals.
For this case, the transfer function of the filter $\left( g_{k_m,k_n} \right)$ is defined as the polynomial
\begin{align*}
	G(z_m,z_n) = \sum_{k_m} \sum_{k_n} \, g_{k_m,k_n} \, z_m^{-k_m} z_n^{-k_n} \text{,}
\end{align*}
where the subscript $m$ refers to the horizontal axis and $n$ to the vertical one.
The $ G^*(z_m,z_n) = G(z_n,z_m) $ is a polynomial transposed to a polynomial $ G(z_m,z_n) $.
In this paper, a shortened notation G is only written in order to keep the notation readable.

A discrete wavelet transform is a mathematical tool which is suitable for the decomposition of a discrete signal into low-pass and high-pass frequency components.
The single-scale transform splits the input signal into two components.
As shown in \cite{Mallat1989}, the transform can be computed by a pair of filters, referred to as $\mathrm{G}_0, \mathrm{G}_1$, followed by subsampling by a factor of two.
Formally, the transform can also be represented by the polyphase matrix
\begin{align}
	\label{eqn:convolution}
	\begin{bmatrix}
		\w{\odd{\mathrm{G}_1}} & \even{\mathrm{G}_1} \\
		\odd{\mathrm{G}_0} & \w{\even{\mathrm{G}_0}}
	\end{bmatrix}
	\text{,}
\end{align}
where the polynomials $\even{\mathrm{G}}$ and $\odd{\mathrm{G}}$ refer to the even and odd terms of $\mathrm{G}$.
This matrix defines the convolution scheme.
Following the instructions by \cite{Daubechies1998}, the convolution scheme in (\ref{eqn:convolution}) can be factored into a sequence
\begin{align}
	\label{eqn:lifting}
	\prod_{K > k > 0}
	\begin{bmatrix}
		1 & \U{}^{(k)} \\
		0 & 1
	\end{bmatrix}
	\begin{bmatrix}
		1 & 0 \\
		\P{}^{(k)} & 1
	\end{bmatrix}
\end{align}
of short filters, known as the lifting scheme.
The filters employed in (\ref{eqn:lifting}) are referred to as the lifting steps.
The first step $\P{}^{(k)}$ in the $k$th pair is referred to as the predict and the second one $\U{}^{(k)}$ to as the update.
The lifting scheme reduces the number of operations by up to half.
Note that the superscript $(k)$ is omitted in the text below.

Considering the GPUs, the processing of single or several transform samples is mapped to independent processing units.
The units must use some sort of synchronization method (barrier) to avoid race conditions.
In the lifting scheme, the barriers are required before the lifting steps.
The barriers are indicated by the $|$ symbol.
For example, $\mathrm{M}_2 | \mathrm{M}_1 $ are two adjacent lifting steps separated by the barrier.

\begin{figure}[h]
	\hspace*{\fill}%
	\subfigure{\includegraphics{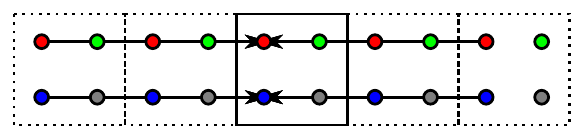}}%
	\hspace*{\fill}\\[-4pt]%
	\hspace*{\fill}%
	\subfigure{\includegraphics{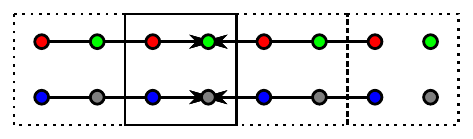}}%
	\hspace*{\fill}%
	\caption{
		Separable convolution for the CDF\,9/7 wavelet (horizontal part only).
		Appropriately chosen pairs of matrix rows are depicted in separate subfigures.
		The arrows denote a multiply--accumulate operation (multiplication by a real constant) and they are pointing to the destination operand.
		Note that some arrows overlap the others.
	}
	\label{fig:dataflow-Separable-Convolution}
\end{figure}

The \mbox{2-D} transform is defined as a tensor product of \mbox{1-D} transforms.
Consequently, the transform splits the signal into a quadruple of wavelet coefficients.
Following the paper of \cite{Mallat1989}, the \mbox{1-D} transforms are applied in both directions sequentially.
By its nature, this scheme can be referred to as the separable convolution.
The calculations in a single direction are performed in a single step (two steps in total).
The scheme can be described as
\begin{align*}
	\NN[]^V \, \big| \, \NN[]^H \, \big| \text{,}
\end{align*}
where $\NN^H$ is \mbox{1-D} transform in horizontal direction and $\NN^V$ is in vertical one.
For the well-known Cohen-Daubechies-Feauveau (CDF) wavelet with 9/7 samples, these matrices are illustrated in \Figure{fig:dataflow-Separable-Convolution} (horizontal part only).
Particularly, the filters in the figure are of sizes 9 and 7 taps.
Note that the color balls correspond to the quadruple of wavelet coefficients.

\begin{figure}[b]
	\hspace*{\fill}%
	\subfigure[${\T[\P]^H}$]{\includegraphics{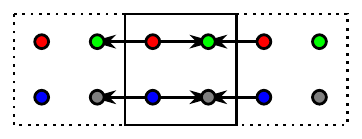}}%
	\hspace*{\fill}
	\subfigure[${\S[\U]^H}$]{\includegraphics{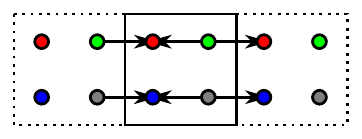}}%
	\hspace*{\fill}%
	\vspace{-2pt}%
	\caption{
		Separable lifting scheme (horizontal part only) for the CDF\,5/3 and 9/7 wavelets.
	}
	\label{fig:dataflow-Separable-Lifting}
\end{figure}

Another scheme widely used for \mbox{2-D} transform is the separable lifting.
Similarly to the convolution, the predict and update lifting steps can be applied in both directions sequentially.
Moreover, horizontal and vertical steps can be arbitrarily interleaved thanks to linear nature of the filters.
In this paper, the scheme is defined as
\begin{align*}
	\S[\U]^V \, \big| \, \S[\U]^H \, \big| \, \T[\P]^V \, \big| \, \T[\P]^H \, \big| \text{,}
\end{align*}
wherein the predict steps $\T$ always precede the update $\S$ ones.
This mapping corresponds to a single $\P$ and $\U$ pair of lifting steps from (\ref{eqn:lifting}).
For multiple pairs, the scheme is separately applied to each such pair.
To describe \mbox{2-D} matrices, the lifting steps must be migrated into two dimensions as $ \+G = G(z_m,z_n) = G(z_m) $.
The individual steps are then defined as
\begin{align*}
	{\T[\P]^H} & =
	\begin{bmatrix}
		\w{1} & 0     & 0     & 0     \\
		\P    & \w{1} & 0     & 0     \\
		0     & 0     & \w{1} & 0     \\
		0     & 0     & \P    & \w{1} \\
	\end{bmatrix} \text{,}
	\\
	{\T[\P]^V} & =
	\begin{bmatrix}
		\w{1} & 0     & 0     & 0     \\
		0     & \w{1} & 0     & 0     \\
		\P^*  & 0     & \w{1} & 0     \\
		0     & \P^*  & 0     & \w{1} \\
	\end{bmatrix} \text{,}
	\\
	{\S[\U]^H} & =
	\begin{bmatrix}
		\w{1} & \U    & 0     & 0     \\
		0     & \w{1} & 0     & 0     \\
		0     & 0     & \w{1} & \U    \\
		0     & 0     & 0     & \w{1} \\
	\end{bmatrix} \text{,}
	\\
	{\S[\U]^V} & =
	\begin{bmatrix}
		\w{1} & 0     & \U^*  & 0     \\
		0     & \w{1} & 0     & \U^*  \\
		0     & 0     & \w{1} & 0     \\
		0     & 0     & 0     & \w{1} \\
	\end{bmatrix} \text{.}
\end{align*}
For the CDF wavelets, the matrices are also illustrated in \Figure{fig:dataflow-Separable-Lifting} (horizontal part only).

Until now, several papers compared the performance of the separable lifting and separable convolution schemes on GPUs.
Exemplarily, the \cite{Tenllado2008} compared these schemes on GPUs using pixel shaders.
The authors mapped data to \mbox{2-D} textures, constituted by four floating-point elements.
They concluded that the separable convolution is more efficient than the separable lifting scheme in most cases.
They further noted that fusing several consecutive kernels might significantly speed up the execution, even if the complexity of the resulting fused pixel program is higher.

To illustrate the problem more widely, also several other papers are discussed.
Kucis \textit{et al.} compared the performance of several recently published schedules (either the convolution or lifting) for computing the \mbox{2-D} DWT using the OpenCL framework.
In more detail, the work compares a convolution-based algorithm \cite{Galiano2011} against several lifting-based methods \cite{Blazewicz2012,Laan2011} in the horizontal part of the transform.
The authors concluded that the lifting-based algorithm of \cite{Blazewicz2012} is the fastest method.
Furthermore, \cite{Laan2011} compared the performance of their separable lifting-based method against a convolution-based method.
They concluded that the lifting is the fastest method.
The authors also compared the performance of implementations in CUDA and pixel shaders, based on the work of \cite{Tenllado2008}.
The CUDA implementation proved to be the faster choice.
In this regard, they noted that a speedup in CUDA occurs because the CUDA effectively makes use of on-chip memory.
This use is not possible in pixel shaders, which exchange the data using off-chip memory.
Without providing further details, other important approaches can be found in the papers of \cite{Matela2009,Galiano2013,Song2014}, and most recently in \cite{Quan2016}.

In our previous works \cite{Barina2016,Kula2016}, we introduced several non-separable schemes for calculation of \mbox{2-D} DWT.
In this paper, we take the most promising scheme and adapt it to particular platforms.
Moreover, differences and similarities between the non-separable scheme and their separable competitors are homogeneously discussed.
All schemes are also analyzed and evaluated.

\section{Proposed Scheme and Optimization}
\label{sec:proposed-schemes-and-improvements}

The existing approaches did not study the possibility of a partial fusion of lifting polyphase matrices.
This section presents an alternative non-separable scheme for the calculation of the \mbox{2-D} DWT.

\begin{figure}[h]
	\vspace{-2pt}%
	\hspace*{\fill}%
	\subfigure[${\T[\P]}$]{\includegraphics{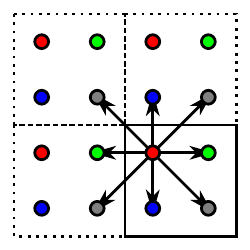}}%
	\hspace*{\fill}%
	\subfigure[${\T[\P]}$]{\includegraphics{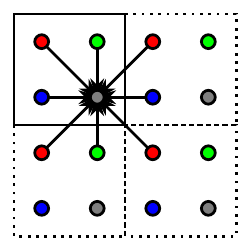}}%
	\hspace*{\fill}\\[-2pt]%
	\hspace*{\fill}
	\subfigure[${\S[\U]}$]{\includegraphics{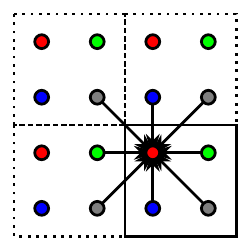}}%
	\hspace*{\fill}%
	\subfigure[${\S[\U]}$]{\includegraphics{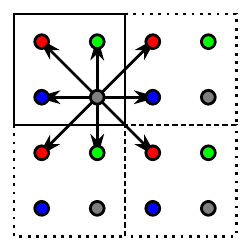}}%
	\hspace*{\fill}%
	\null%
	\vspace{-2pt}%
	\caption{
		Non-separable lifting scheme for the CDF wavelets.
	}%
	\vspace{-2pt}%
	\label{fig:dataflow-Non-Separable-Lifting}
\end{figure}

By combining the corresponding horizontal and vertical steps of the separable lifting scheme, the non-separable lifting scheme is formed.
The number of operations has slightly been increased.
The scheme consists of a spatial predict and spatial update step, thus two steps in total for each pair of the original lifting steps.
For each pair of $\P$ and $\U$, the scheme follows from
\begin{align*}
	\S[\U] \, \big| \, \T[\P] \, \big| \text{,}
\end{align*}
where
\begin{align*}
	{\T[\P]} & =
	\begin{bmatrix}
		\w{1}  & 0     & 0     & 0     \\
		\P     & \w{1} & 0     & 0     \\
		\P^*   & 0     & \w{1} & 0     \\
		\P\P^* & \P^*  & \P    & \w{1} \\
	\end{bmatrix} \text{,}
	\\
	{\S[\U]} & =
	\begin{bmatrix}
		\w{1} & \U    & \U^*  & \U\U^* \\
		0     & \w{1} & 0     & \U^*   \\
		0     & 0     & \w{1} & \U     \\
		0     & 0     &     0 & \w{1}  \\
	\end{bmatrix} \text{.}
\end{align*}
Note that the spatial filters in $\P\P^*$ and $\U\U^*$ may be computationally demanding, depending on their sizes.
For the CDF wavelets, the scheme is graphically illustrated in \Figure{fig:dataflow-Non-Separable-Lifting}.

An important observation can be made regardless the underlying platform.
A very special form of the operations guarantees that the processing units never access the results belonging to their neighbors.
These operations comprise only constants. 
Since the convolution is a linear operation, the polynomials can be collected out from the original matrices, and calculated in a different step.
The original polynomials are split as $\P = \P[0] + \P[1]$ and $\U = \U[0] + \U[1]$.
The $\P[0]$ and $\U[0]$ are the constants. 
As a next step, the $\P[0]$ and $\U[0]$ are substituted into the separable lifting scheme.
On the contrary, the $\P[1]$ and $\U[1]$ are kept in the original non-separable lifting scheme.
These two steps are then computed without any barrier.
The observation is exploited to adapt the scheme for a particular platform.

The schemes for the shaders and OpenCL exploit the above-described observation with $\P[0]$, $\U[0]$ polynomials.
Implementations in the shaders map input and output data to \mbox{2-D} textures.
There is no possibility to retain results in registers, and the results are exchanged through textures in off-chip memory.
Considering the OpenCL implementations, the image is divided into overlapping blocks and on-chip memory shared by all threads in a block is utilized to exchange the results.
Additionally, some results are passed in registers.
\vspace{-2pt}%

\begin{table}[h!]
	\vspace{-6pt}%
	\caption{%
		Steps and arithmetic operations for the optimized schemes.%
	}
	\vspace{-2pt}%
	\subtable[CDF\,5/3]{%
		\begin{tabu} to \linewidth {X[1.5r]X[l]X[c]X[c]X[c]}
		\toprule
			\multicolumn{2}{c}{\qquad\qquad{}scheme} & steps & \multicolumn{2}{c}{operations} \\
			            ~ &               ~ &     ~ & \multicolumn{1}{c}{OpenCL} & \multicolumn{1}{c}{shaders} \\
		\midrule
			separable     & convolution     &  2 &  20 &  22 \\
			separable     & lifting         &  4 &  16 &  16 \\
		\midrule
			non\nobreakdash-separable & lifting         &  2 &  18 &  18 \\
		\bottomrule
		\end{tabu}%
	}\\[-4pt]%
	\subtable[CDF\,9/7]{%
		\begin{tabu} to \linewidth {X[1.5r]X[l]X[c]X[c]X[c]}
		\toprule
			\multicolumn{2}{c}{\qquad\qquad{}scheme} & steps & \multicolumn{2}{c}{operations} \\
			            ~ &               ~ &     ~ & \multicolumn{1}{c}{OpenCL} & \multicolumn{1}{c}{shaders} \\
		\midrule
			separable     & convolution     &  2 &  56 &  58 \\
			separable     & lifting         &  8 &  32 &  32 \\
		\midrule
			non\nobreakdash-separable & lifting         &  4 &  36 &  36 \\
		\bottomrule
		\end{tabu}%
	}\\[-4pt]
	\label{tab:parameters-baseline}
\end{table}

This paper explores the performance for two frequently used wavelets, namely, CDF\,5/3 and CDF\,9/7 \cite{Cohen1992}.
Their fundamental properties are listed in \Table{tab:parameters-baseline}.
Note that the number of operations is commonly proportional to a transform performance.
Additionally, the number of steps correspond to the number of synchronizations on parallel architectures, which also form a performance bottleneck.

\section{Evaluation}
\label{sec:performance}

The experiments were performed on GPUs of two biggest vendors using the OpenCL and pixel shaders.
Only a transform performance was measured, usually in GB/s (gigabytes per second).
The host system does not help in the calculation.
Only results for two cards (AMD Radeon HD 6970, NVIDIA Titan X) are shown due to limited space.
Their technical parameters are summarized in \Table{tab:gpus}.

\begin{table}[b]
	\vspace{-12pt}%
	\caption{
		Description of the GPUs used for the evaluation.
	}%
	\begin{tabu} to \linewidth {l|X[r]X[r]}
		\toprule
		label            &       AMD 6970 &   NVIDIA Titan X \\
		\midrule
		model            & Radeon HD 6970 & Titan X (Pascal) \\
		\midrule
		multiprocessors  &             24 &               28 \\
		total processors &         1\,536 &           3\,584 \\
		performance      & 2\,703\,GFLOPS &  10\,157\,GFLOPS \\
		\midrule
		bandwidth        &      176\,GB/s &        480\,GB/s \\
		on-chip memory   &        32\,KiB &          96\,KiB \\
		\bottomrule
	\end{tabu}
	\label{tab:gpus}
\end{table}

\begin{figure}[h!]
	\centering
	\subfigure[CDF\,5/3 in OpenCL]{\includegraphics[width=.9\linewidth]{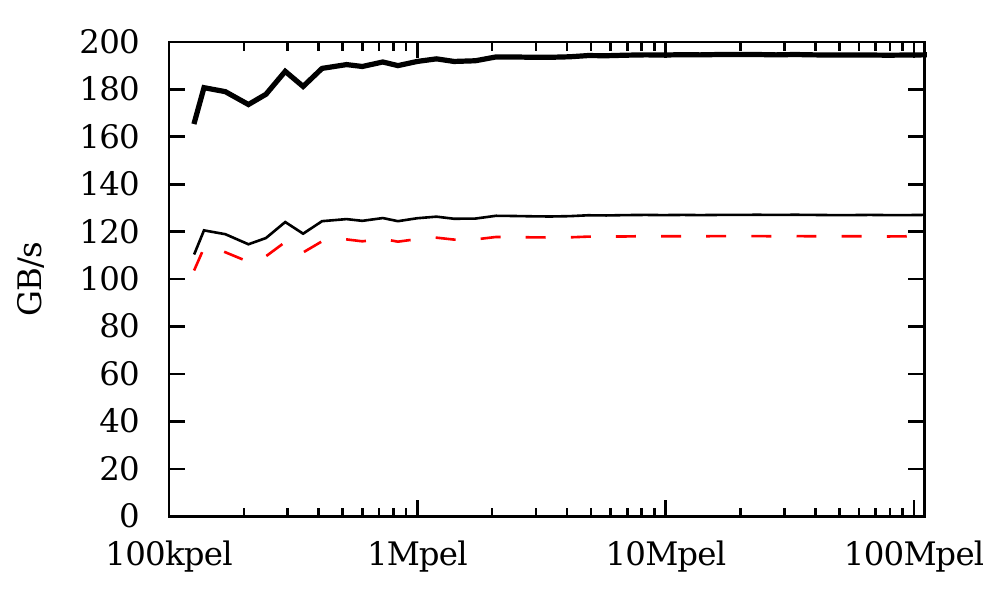}}\\[-2pt]%
	\subfigure[CDF\,5/3 in pixel shader]{\includegraphics[width=.9\linewidth]{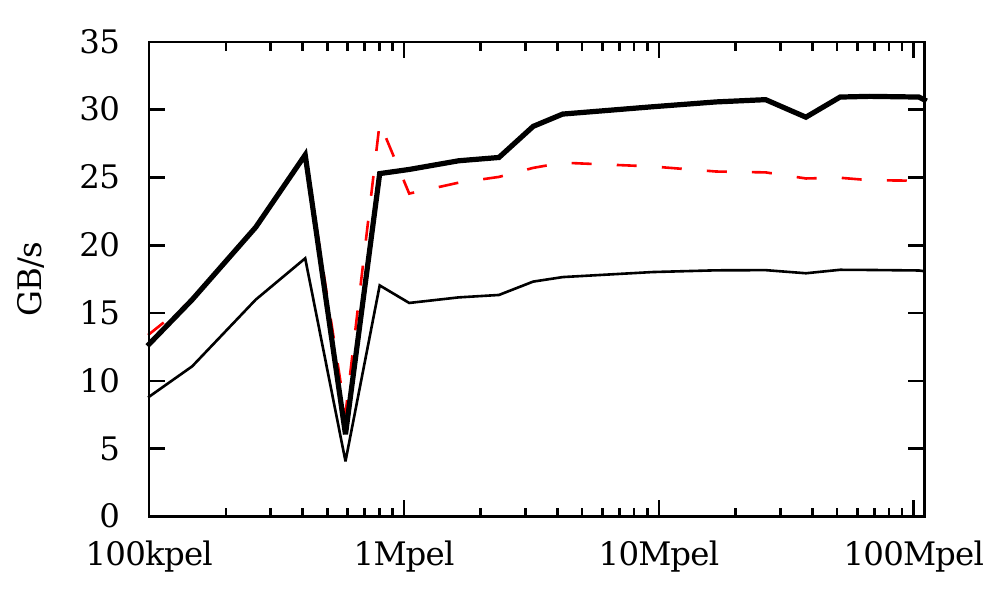}}\\[-2pt]%
	\subfigure[CDF\,9/7 in OpenCL]{\includegraphics[width=.9\linewidth]{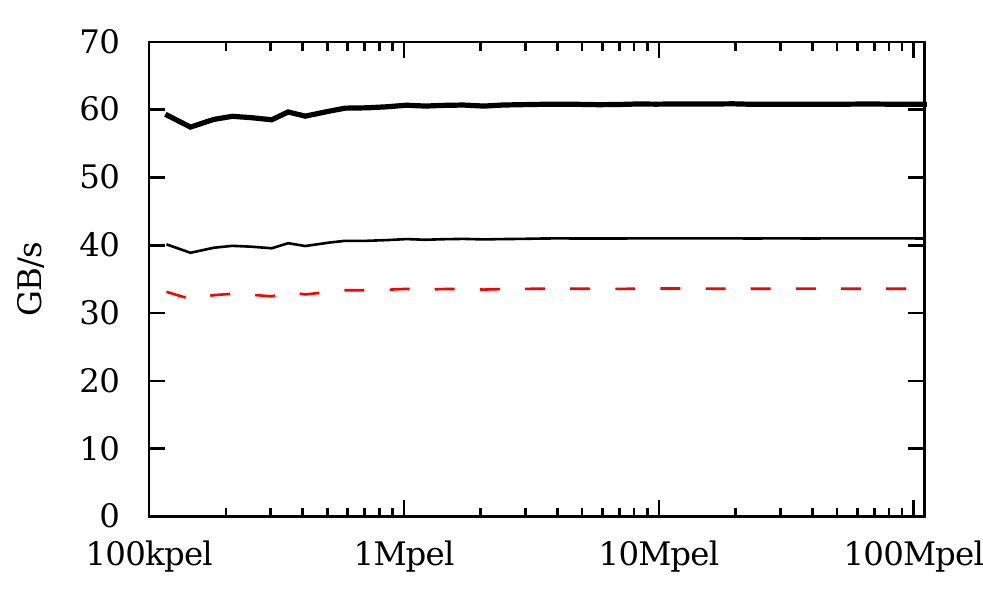}}\\[-2pt]%
	\subfigure[CDF\,9/7 in pixel shader]{\includegraphics[width=.9\linewidth]{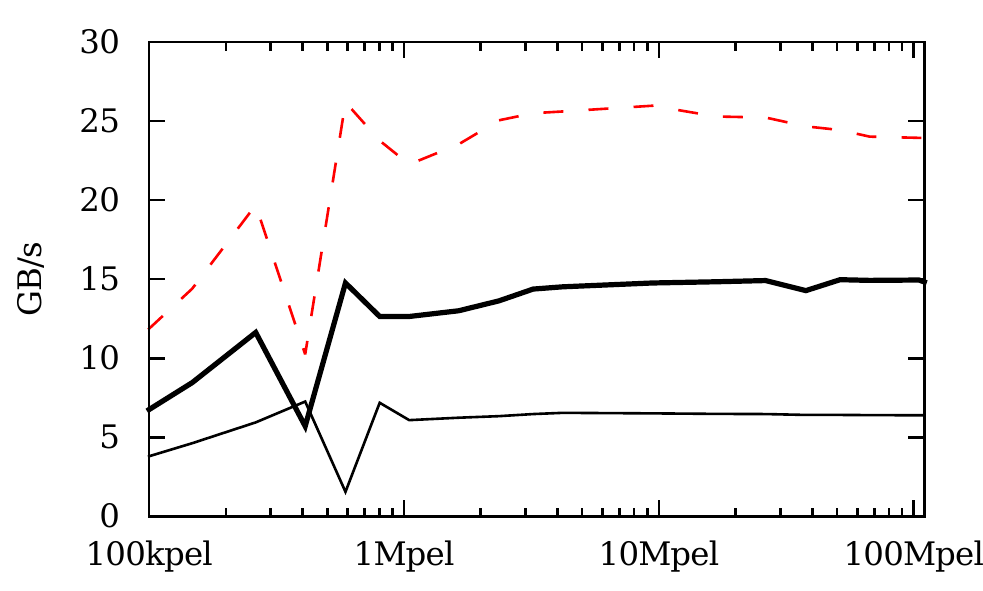}}\\[-2pt]%
	\includegraphics[width=\linewidth]{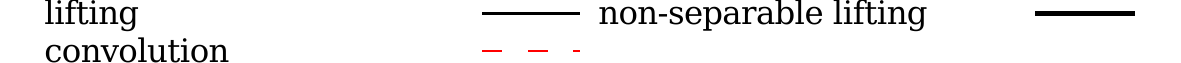}%
	\caption{
		Performance (gigabytes per second) for both CDF wavelets.
	}
	\label{fig:plots}
\end{figure}

The implementations were created using the DirectX HLSL and OpenCL.
The HLSL implementation is used on the NVIDIA Titan X, whereas the OpenCL implementation on the AMD 6970.
The results of the performance comparison are shown in \Figure{fig:plots}.
The non-separable lifting schemes always outperform the separable lifting.
Similar results were also achieved on other cards, especially considering the shaders.
Looking at the experiments with the pixel shaders, some transients can be seen at the beginning of the plots.
We concluded that these transients are caused by a suboptimal use of cache system, or alternatively by some overhead made by the graphics API.

\section{Conclusions}
\label{sec:conclusion}

This paper introduced and discussed non-separable lifting scheme for computation of the \mbox{2-D} discrete wavelet transform on modern GPUs.
As an option, an optimization strategy leading to a reduction in the number of arithmetic operations was presented.
Using this strategy, the introduced scheme was adapted on the OpenCL framework and pixel shaders.
The implementations were then evaluated using GPUs of two biggest vendors.
At least for the CDF wavelets, the non-separable lifting scheme exhibit mostly a better performance than their separable competitors.

We reached the following conclusion.
Fusing several consecutive steps of the schemes might significantly speed up the execution, irrespective of their higher complexity.
Note that all of the schemes are general and they can be used on any discrete wavelet transform.
In future work, we plan to tackle with multi-scale transforms.

\bibliographystyle{IEEEtran}
\bibliography{IEEEabrv,sources}

\end{document}